\documentclass[letterpaper, 10 pt, conference]{ieeeconf}  % Comment this line out if you need a4paper

\IEEEoverridecommandlockouts                              % This command is only needed if
\usepackage{graphicx} % for pdf, bitmapped graphics files
\graphicspath{{figures/}}

\usepackage{amsthm}
\usepackage{amsmath} % assumes amsmath package installed
\usepackage{amssymb}  % assumes amsmath package installed
\usepackage{mathtools}
\usepackage{xcolor} % for extra color names
\usepackage{ulem} % for strike through command \sout
\usepackage{multirow}
\usepackage{color}
\usepackage{latexsym}
\usepackage{multicol}

\usepackage[font=small,labelfont=bf]{caption}
\usepackage{lipsum}
\usepackage{subcaption}



\renewcommand{\thefigure}{\arabic{figure} }

\newtheorem{definition}{Definition}

\newcommand*{\mydprime}{^{\prime\prime}\mkern-1.2mu}

\usepackage[subnum]{cases}

\newcommand{\Next}{\bigcirc}
\newcommand{\Always}{\Box}
\newcommand{\Event}{\diamondsuit}
\newcommand{\Until}{\mathcal{U}}
\newcommand{\Implies}{\Rightarrow}
\newcommand{\Then}{\mathcal{T}}
\newcommand{\True}{\top}

\newcommand{\ignore}[1]{%
}

\title{Reinforcement Learning With Temporal Logic Rewards
}

\author{Xiao Li, Cristian-Ioan Vasile and Calin Belta% <-this % stops a space
\thanks{X. Li and C. Belta are with Boston University, Boston, MA. Email: \{xli87,cbelta\}@bu.edu. C.-I. Vasile is with Massachusetts Institute of Technology, Cambridge, MA. Email: cvasile@mit.edu }.
\thanks{This work is partially supported by the ONR under grants N00014-14-1-0554 and by the NSF under grants NRI-1426907 and CMMI-1400167}}

\begin{document}

%\author{\authorblockN{Michael Shell}
%\authorblockA{School of Electrical and\\Computer Engineering\\
%Georgia Institute of Technology\\
%Atlanta, Georgia 30332--0250\\
%Email: mshell@ece.gatech.edu}
%\and
%\authorblockN{Homer Simpson}
%\authorblockA{Twentieth Century Fox\\
%Springfield, USA\\
%Email: homer@thesimpsons.com}
%\and
%\authorblockN{James Kirk\\ and Montgomery Scott}
%\authorblockA{Starfleet Academy\\
%San Francisco, California 96678-2391\\
%Telephone: (800) 555--1212\\
%Fax: (888) 555--1212}}

% avoiding spaces at the end of the author lines is not a problem with
% conference papers because we don't use \thanks or \IEEEmembership

% for over three affiliations, or if they all won't fit within the width
% of the page, use this alternative format:
%
%\author{\authorblockN{Michael Shell\authorrefmark{1},
%Homer Simpson\authorrefmark{2},
%James Kirk\authorrefmark{3},
%Montgomery Scott\authorrefmark{3} and
%Eldon Tyrell\authorrefmark{4}}
%\authorblockA{\authorrefmark{1}School of Electrical and Computer Engineering\\
%Georgia Institute of Technology,
%Atlanta, Georgia 30332--0250\\ Email: mshell@ece.gatech.edu}
%\authorblockA{\authorrefmark{2}Twentieth Century Fox, Springfield, USA\\
%Email: homer@thesimpsons.com}
%\authorblockA{\authorrefmark{3}Starfleet Academy, San Francisco, California 96678-2391\\
%Telephone: (800) 555--1212, Fax: (888) 555--1212}
%\authorblockA{\authorrefmark{4}Tyrell Inc., 123 Replicant Street, Los Angeles, California 90210--4321}}

\normalem % delete this if you don't use ``ulem'' package

\maketitle

\begin{abstract}
\emph{Reinforcement learning} (RL) depends critically on the choice of reward functions used to capture
the desired behavior and constraints of a robot.
Usually, these are handcrafted by a expert designer and represent heuristics for relatively simple tasks.
Real world applications typically involve more complex tasks with rich temporal and logical structure.
In this paper we take advantage of the expressive power of {\em temporal logic (TL)} to specify complex rules
the robot should follow, and incorporate domain knowledge into learning.
We propose {\em Truncated Linear Temporal Logic} (TLTL) as specifications language,
that is arguably well suited for the robotics applications,
together with quantitative semantics, i.e., {\em robustness degree}.
We propose a RL approach to learn tasks expressed as TLTL formulae that uses
their associated robustness degree as reward functions,
instead of the manually crafted heuristics trying to capture the same specifications.
We show in simulated trials that learning is faster and policies obtained using the proposed approach
outperform the ones learned using heuristic rewards in terms of the robustness degree,
i.e., how well the tasks are satisfied.
Furthermore, we demonstrate the proposed RL approach in a toast-placing task learned by a Baxter robot.
%
%The reward function plays a critical role in {\em reinforcement learning} (RL). It is a place where designers specify the desired behavior and impose important constraints for the system. While most reward functions used in the current RL literature have been  based on heuristics for relatively simple tasks, real world applications typically involve tasks that are logically more complex. In this paper we take advantage of the expressive power of {\em temporal logic (TL)} to express complex rules the system should follow, or incorporate domain knowledge into learning. We propose the \textit{Truncated Linear Temporal Logic} (TLTL) which we argue is suitable for robotic task specification. We show in simulated manipulation tasks that using TLTL specifications and their robustness as rewards result in faster learning and better policies compared to rewards designed from heuristic. We demonstrate the use of TL in RL in a toast-placing task learned by a Baxter robot.
\end{abstract}

\IEEEpeerreviewmaketitle

\section{Introduction}
\label{sec:1}

Reinforcement learning methods have the ability of finding well behaved
controllers (or policies)%
%\footnote{Used interchangeably throughout the paper.}
for robots without the need to know their internal structure and dynamical details,
and the environment they operate in.
An often overlooked issue in reinforcement learning research is the design of
effective reward functions.

Real world applications usually involve complex tasks that may not be defined
as reach-avoid operation (``go from A to B while avoiding obstacles'').
Consider the task of driving a car for a humanoid robot.
A simple extrinsic reward function is the travel duration to destination.
However, the robot will need a unacceptable large number of trials
to learn to apply the gas and brake, use the steering wheel and transmission,
and drive safe, using just the extrinsic reward.
On the other hand, humans have mastered driving to an acceptable level
of proficiency, and general rules have been defined to perform and learn
the skill faster and more reliable.
Formally incorporating known rules in the reward function dramatically accelerates
the time to learn new skills, where
correct behaviors (e.g., always put at least one hand on the steering wheel) are encouraged,
and hazardous behaviors (e.g., step on the gas and brake pedal at the same time)
are penalized.

%Specifying a task in real world scenarios is usually not as simple as
%``going from A to B while avoiding obstacles''.
%Suppose we want a humanoid robot to learn how to drive a car.
%An obvious extrinsic reward function can be defined
%proportional to the forward moving speed.
%But asking the robot to learn to apply gas and brake, use the steering wheel, and transmission
%in the correct way with only the extrinsic reward will take an unacceptable large number of trials.
%On the other hand, humans have well mastered the skill of driving,
%and it is not difficult to come up with a set of rules for driving a vehicle.
%Formally incorporating known rules in the reward function so that correct behaviors
%(e.g., always put at least one hand on the steering wheel) are encouraged,
%and hazardous behaviors (e.g., step on the gas and brake pedal at the same time)
%are penalized will dramatically accelerate the time to learn new skills.

\begin{figure}[tbh]
\label{fig:1}
\begin{center}
\includegraphics[width=0.65\linewidth]{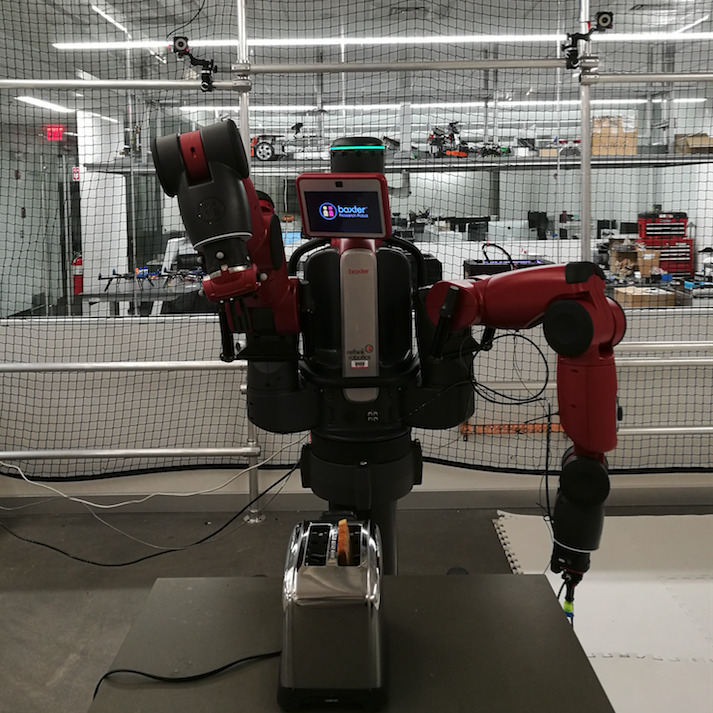}
\caption{Toast-placing task learned using temporal logic rewards.
Our framework allows learning of temporally structured tasks using simple policy search algorithms.}
\end{center}
\end{figure}

The problem of accurately incorporating complex specifications in reward functions
is referred to in the literature as {\em reward hacking} in~\cite{Amodei2016}.
The inability of ad-hoc rewards to capture the semantics of complex tasks has negative
repercussions on the learned policies.
Policies that maximize the reward functions are not guaranteed to satisfy the specifications.
Furthermore, it is not easy to design and prove that increasing rewards translate
to better satisfaction of the specifications.
%
%Not being able to expressively and accurately incorporate complex specifications
%in the reward function has another negative implication, which is referred to as
%{\em reward hacking} in~\cite{Amodei2016}.
%The goal of a RL agent is to find a policy that maximizes the obtained reward.
%Depending on how the reward function is defined, the resultant policy may not coincide
%with the designer's expectations.
%
A simple example from~\cite{Amodei2016} that highlights these problems
involves a robot learning to clean an office.
%
%is a robot learning to clean an office as described in~\cite{Amodei2016}.
If only a positive reward is given when the robot cleans up a mess
(picks up trash from the ground),
then the robot may learn to first make a mess and then clean it up.
Imperfect reward functions provide opportunities for a learning robot to exploit, and
find high gain solutions that are algorithmically correct, but deviates from the designer's
intentions.

In this paper, we use formal specification languages to capture the designer's requirements
of what the robot should achieve.
We propose {\em Truncated Linear Temporal Logic (TLTL)} as a specification language
with an extended set of operators defined over finite-time trajectories of a robot's states.
TLTL provides convenient and effective means of incorporating complex intentions,
domain knowledge, and constraints into the learning process.
We define quantitative semantics (also referred to as {\em robustness degree}) for TLTL.
The robustness degree is used to transform temporal logical formulae into real-valued reward functions.

We compare the convergence rate and the quality of learned policies of RL algorithms using
temporal logic (i.e., robustness degree) and heuristic reward functions.
In addition, we compare the results of a simple TL algorithm against a more elaborate RL algorithm
with heuristic rewards.
In both cases better quality policies were learned faster using the proposed approach with TL rewards
than the heuristic reward functions.
%
%We show that in both cases better quality policies are learned from TL reward in comparable or shorter time than those learned from a heuristic reward.

\section{Background}
We will use the terms controller and policy interchangeably throughout the paper.

\subsection{Policy Search in Reinforcement Learning}

In this section we briefly introduce a class of reinforcement learning methods called policy search methods.
Policy search methods have exhibited much potential in finding satisfactory policies in Markov Decision Processes (MDP) with continuous state and action spaces (also referred to as an infinite MDP), which especially suits the need for finding a controller in robotic applications \cite{Deisenroth2011}.

\begin{definition}
\label{def:1}
An infinite MDP is a tuple $\langle S,A,p(\cdot,\cdot,\cdot),R(\cdot)\rangle$, where $S \subseteq {\rm I\!R}^n$ is a continuous set of states; $A \subseteq {\rm I\!R}^m$ is a continuous set of actions; $p: S \times A \times S \to [0,1]$ is the transition probability function with $p(s,a,s^{\prime})$ being the probability of taking action $a \in A$ at state $s \in S$ and ending up in state $s^{\prime} \in S$ (also commonly written as a conditional probability $p(s^\prime | s, a)$); $R: \tau \to {\rm I\!R}$ is a reward function where $\tau = (s_0, a_0, ..., s_T, )$ is the state-action trajectory, $T$ is the horizon.
\end{definition}
In RL, the transition function $p(s,a,s^\prime)$ is unknown to the learning agent. The reward function $R(\tau)$ can be designed or learn (as in the case of inverse reinforcement learning). The goal of RL is to find an optimal stochastic policy $\pi^\star: S \times A \to [0,1]$ that maximizes the expected accumulated reward, i.e.
\begin{equation}
\label{eq:2}
\pi^\star = \underset{\pi}{\arg\max} \, E_{p^\pi(\tau)}\left[ R(\tau)\right],
\end{equation}
$p^\pi(\tau)$ is the trajectory distribution from following policy $\pi$. And $R(\tau)$ is the reward obtained given $\tau$.

In policy search methods, the policy is represented by a parameterized model (e.g., neural network, radial basis function) denoted by $\pi(s,a|\theta)$ (also written as $\pi_\theta$ in short) where $\theta$ is the set of model parameters. Search is then conducted in the policy's parameter space to find the optimal set of $\theta$ that achieves~\eqref{eq:2}

\begin{equation}
\label{eq:3}
\theta^\star = \underset{\theta}{\arg\max} \, E_{p^{\pi_\theta}(\tau)}\left[ R(\tau)\right],
\end{equation}

Many policy search methods exist to solve the above problem.
The authors of~\cite{Deisenroth2011} provide a survey on policy search methods
applied in robotics.
In this work we adopt the Relative Entropy Policy Search (REPS) technique.
A brief overview of the method is given in the next section.

\subsection{Relative Entropy Policy Search}

Relative Entropy Policy Search is an information-theoretic approach that solves
the policy search problem.
The episode-based version of REPS can be formulated as the following constrained
optimization problem

\begin{equation}
\label{eq:21}
\underset{p(\tau)}{\max} \, E_{p(\tau)}\left[ R(\tau)\right]
\,\, s.t.  \,\,D_{KL}(p(\tau) || q^{\pi_\theta}(\tau)) < \epsilon,
\end{equation}

\noindent where $q^{\pi_\theta}(\tau)$ is the trajectory distribution following the existing policy. $D_{KL}()$ is the KL-divergence between two policies and $\epsilon$ is a threshold. This constraint limits the step size a policy update can take and ensures that the trajectory distribution resulting from the updated policy stays near the sampled trajectories. This constraint not only promotes exploration safety which is especially important in robotic applications, it also helps the agent to avoid premature convergence.

The optimization problem in~\eqref{eq:21} can be solved using the Lagrange multipliers method which results in a closed-form trajectory distribution update equation given by
\begin{equation}
p(\tau) \propto q(\tau)\exp \left(\frac{R(\tau)}{\eta}\right).
\end{equation}
Since we only have sample trajectories ($\tau_{1,...,N}$), we can estimate $p(\tau)$ only at the sampled points by

\begin{equation}
p(\tau_i) \propto \exp \left(\frac{R(\tau_i)}{\eta}\right).
\end{equation}

\noindent $q(\tau_i)$ is dropped from the above result because we are already sampling from $q(\tau)$. $\eta$ is the Lagrange multiplier obtain from optimizing the dual function

\begin{equation}
g(\eta) = \eta\epsilon + \eta\log\sum_i\frac{1}{N}\exp\left(\frac{R(\tau_i)}{\eta}\right)
\end{equation}
\noindent We refer interested readers to \cite{Deisenroth2011} for detailed derivations.

We adopt the time-varying linear-Gaussian policies $\pi_{\theta_t} = \mathcal{N}(K_ts_t+k_t, \Sigma_t)$ (here $\theta_t=( k_t, \Sigma_t)$ for $t=0,...,T$)
and weighted maximum-likelihood estimation to update the policy parameters (feedback gain $K_t$ is kept fixed to reduce the dimension of the parameter space).
This approach has been used in~\cite{chebotar2016path}. The difference is that \cite{chebotar2016path} recomputes $p(\tau_i)$ at each step $t$ using cost-to-go before updating $\theta_i$. Since a temporal logic reward (described in the next section) depends on the entire trajectory, it doesn't have the notion of cost-to-go and can only be evaluated as a terminal reward. Therefore $p(\tau_i)$ (written short as $p_i$) is computed once and used for updates of all $\theta_t$ (similar approach used in episodic PI-REPS \cite{gomez2014policy}). The resulting update equations are

\begin{equation}
\begin{split}
&k_t^\prime = \sum_i^N p_i k_{i,t} \\
&\Sigma_t^\prime = \sum_i^N p_i (k_{i,t} - k_t^\prime)(k_{i,t} - k_t^\prime)^T,
\end{split}
\end{equation}

\noindent where $k_{i,t}$ is the feed-forward term in the time-varying linear-Gaussian policy at time $t$ and for sample trajectory $i$. 

\section{Truncated Linear Temporal Logic(TLTL)}
\label{sec:3}
In this section, \textit{we propose TLTL, a new temporal logic that we argue is well suited for specifying goals and introducing domain knowledge for the RL problem.} In the following definitions, the sets of real and integer numbers are denoted by IR and Z, respectively. The subset of integer numbers {a,...,b 1}, a,b 2 Z, a < b, is denoted by [a, b), and [a, b] = [a, b) [ {b}.

\subsection{TLTL Syntax And Semantics}
\label{sec:3a}
A TLTL formula is defined over predicates of form $f(s) < c$,
where $f: {\rm I\!R}^n \rightarrow {\rm I\!R}$ is a function and $c$ is a constant. A TLTL specification has the following syntax:
\begin{equation}
\begin{split}
\phi := \ & \True \,\,|\,\, f(s) < c \,\,| \,\, \neg \phi \,\,|\,\, \phi \wedge \psi \,\,|\,\, \phi \vee \psi \,\,|\,\, \\
        & \Event \phi \,\,|\,\, \Always \phi \,\,|\,\, \phi \, \Until \, \psi \,\,|\,\, \phi\, \Then\, \psi \,\,|\,\, \Next \phi \,\,|\,\, \phi \Implies \psi,
\end{split}
\end{equation}
where $f(s) < c$ is a predicate,
$\neg$~(negation/not), $\wedge$~(conjunction/and), and $\vee$~(disjunction/or) are
Boolean connectives,
and $\Event$~(eventually), $\Always$~(always), $\Until$~(until), $\Then$~(then), $\Next$~(next),
are temporal operators.
Implication is denoted by $\Implies$~(implication). TLTL formulas are evaluated against finite time sequences over a set $S$. As it will become clear later, such sequences will be produced by a Markov Decision Process (MDP, see Definition~\ref{def:1}).

We denote $s_t \in S$ to be the state at time $t$, and $s_{t:t+k}$ to be a sequence of states
(state trajectory) from time $t$ to $t+k$, i.e., $s_{t:t+k}=s_ts_{t+1}...s_{t+k}$. The Boolean semantics of TLTL is defined as:
\begin{alignat*}{3}
&s_{t:t+k} \models f(s)<c \quad &&\Leftrightarrow \quad &&f(s_t) <c, \\
&s_{t:t+k} \models \neg \phi \quad &&\Leftrightarrow \quad &&\neg(s_{t:t+k}\models \phi),\\
&s_{t:t+k} \models \phi \Rightarrow \psi  \quad &&\Leftrightarrow \quad && (s_{t:t+k} \models \phi) \Rightarrow (s_{t:t+k} \models \psi),\\
&s_{t:t+k} \models \phi \wedge \psi \quad &&\Leftrightarrow \quad && (s_{t:t+k} \models \phi) \wedge (s_{t:t+k} \models \psi),\\
&s_{t:t+k} \models \phi \vee \psi \quad &&\Leftrightarrow \quad && (s_{t:t+k} \models \phi) \vee (s_{t:t+k} \models \psi),\\
&s_{t:t+k} \models \Next \phi  \quad &&\Leftrightarrow \quad && (s_{t+1:t+k} \models \phi) \wedge (k>0), \\
&s_{t:t+k} \models \Always \phi \quad &&\Leftrightarrow \quad && \forall t^\prime \in [t,t+k) \  s_{t^\prime:t+k} \models \phi,\\
&s_{t:t+k} \models \Event \phi \quad &&\Leftrightarrow \quad && \exists t^\prime \in [t,t+k) \ s_{t^\prime:t+k} \models \phi,\\
&s_{t:t+k} \models \phi \,\, \Until \,\, \psi \quad &&\Leftrightarrow \quad && \exists t^\prime \in [t,t+k) \,\,s.t.\,\, s_{t^\prime:t+k} \models \psi \\
&\,&&\,&& \wedge (\forall t^{\prime\prime} \in [t,t^\prime) \ s_{t^{\prime\prime}:t^\prime} \models \phi),\\
&s_{t:t+k} \models \phi \,\, \mathcal{T}\,\, \psi \quad &&\Leftrightarrow \quad && \exists t^\prime \in [t,t+k) \,\,s.t.\,\, s_{t^\prime:t+k} \models \psi \\
&\,&&\,&& \wedge (\exists t^{\prime\prime} \in [t,t^\prime) \ s_{t^{\prime\prime}:t^\prime} \models \phi).
\end{alignat*}
Intuitively, state trajectory $s_{t:t+k}\models \Always \phi$ if the specification defined by
$\phi$ is satisfied for every subtrajectory $s_{t^\prime:t+k},\,\,t^\prime \in [t,t+k)$.
Similarly, $s_{t:t+k}\models \Event \phi$ if $\phi$ is satisfied for at least one subtrajectory
$s_{t^\prime:t+k},\,\,t^\prime \in [t,t+k)$.
$s_{t:t+k}\models \phi \,\, \Until \,\, \psi$ if $\phi$ is satisfied at every time step before
$\psi$ is satisfied, and $\psi$ is satisfied at a time between $t$ and $t+k$.
$s_{t:t+k}\models \phi \,\, \Then \,\, \psi$ if $\phi$ is satisfied at least once before
$\psi$ is satisfied between $t$ and $t+k$.
A trajectory $s$ of duration $k$ is said to satisfy formula $\phi$ if $s_{0:k} \models \phi$.

We equip TLTL with quantitative semantics (robustness degree)
, i.e., a real-valued function $\rho(s_{t:t+k}, \phi)$ of state trajectory $s_{t:t+k}$
and a TLTL specification $\phi$ that indicates how far $s_{t:t+k}$ is from satisfying or
violating the specification $\phi$.
The quantitative semantics of TLTL is defined as follows:
\begin{alignat*}{3}
&\rho(s_{t:t+k}, \True)\quad && = \quad && \rho_{max},\\
&\rho(s_{t:t+k},f(s_t)<c) \quad && = \quad &&c-f(s_t),\\
&\rho(s_{t:t+k},\neg \phi) \quad && = \quad &&-\rho(s_{t:t+k},\phi),\\
&\rho(s_{t:t+k}, \phi \,\, \Rightarrow \psi) \quad && = \quad && \max(-\rho(s_{t:t+k}, \phi), \rho(s_{t:t+k}, \psi))\\
&\rho(s_{t:t+k},\phi_1\wedge \phi_2) \quad &&= \quad &&\min(\rho(s_{t:t+k},\phi_1),\rho(s_{t:t+k},\phi_2)), \\
&\rho(s_{t:t+k},\phi_1\vee \phi_2) \quad &&= \quad &&\max(\rho(s_{t:t+k},\phi_1),\rho(s_{t:t+k},\phi_2)),\\
&\rho(s_{t:t+k}, \Next \phi) \quad && = \quad && \rho(s_{t+1:t+k},\phi) \,\,(k>0), \\
&\rho(s_{t:t+k},\Always \phi) \quad &&= \quad && \underset{t^{\prime} \in [t,t+k)}{\min}(\rho(s_{t^{\prime}:t+k},\phi)),\\
&\rho(s_{t:t+k},\Event \phi) \quad &&= \quad && \underset{t^{\prime} \in [t,t+k)}{\max}(\rho(s_{t^{\prime}:t+k},\phi)),\\
&\rho(s_{t:t+k},\phi \,\, \Until \,\, \psi) \quad && = \quad && \underset{t^{\prime} \in [t,t+k)}{\max}( \min (\rho(s_{t^{\prime}:t+k},\psi), \\
& \, &&\, && \underset{t^{\mydprime} \in [t,t^{\prime})}{\min}\rho(s_{t\mydprime:t^\prime},\phi))),\\
&\rho(s_{t:t+k},\phi \,\, \Then \,\, \psi) \quad && = \quad && \underset{t^{\prime} \in [t,t+k)}{\max}( \min (\rho(s_{t^{\prime}:t+k},\psi), \\
& \, &&\, && \underset{t^{\mydprime} \in [t,t^{\prime})}{\max}\rho(s_{t\mydprime:t^\prime},\phi))),
\end{alignat*}
where $\rho_{max}$ represents the maximum robustness value.
Moreover, $\rho(s_{t:t+k},\phi) > 0 \Rightarrow s_{t:t+k} \models \phi$ and
$\rho(s_{t:t+k},\phi) < 0 \Rightarrow s_{t:t+k} \not\models \phi$,
which implies that the robustness degree can substitute Boolean semantics in order to enforce
the specification $\phi$. As an example, consider
specification $\phi = \diamondsuit(s < 10)$, where $s$ is a one dimensional state,
and a two step state trajectory $s_{0:2}=s_0s_1=[11,5]$.
The robustness is
$\rho(s_{0:1}, \phi) = \underset{t \in \{0,1\}}{\max}(10 - s_t) = \max(-1,5) = 5$. Since $\rho(s_t, \phi) > 0$, $s_{0:1} \models \phi$ and the value $\rho(s_t, \phi) = 5$ is a measure of the satisfaction margin (refer to \textit{Example 1} in \cite{Aksaray2016} for a more detail example on task specification using TL and robustness).

\subsection{Comparison With Existing Formal Languages}
In our view, a formal language for RL task specification should have the following characteristics: (1) The language should be defined over predicates so tasks can be conveniently specified as functions of states (2) The language should provide quantitative semantics as a continous measure of its satisfaction. (3) The specification formula should be evaluated over finite sequences (state trajectories) of variable length, thus allow for per-step evaluation on currently available data. (4) Temporal operators can have time bounds but should not require them. A wide variety of formal specification languages exists and some possess parts of the above characteristics. we selectively analyze three specification languages, namely Signal Temporal Logic(STL)~\cite{STL1} (related to Metric Temporal Logic (MTL), omitted here for simplicity), Bounded Temporal Logic (BLTL)~\cite{Latvala2004} and Linear Temporal Logic on Finite Traces ($\textrm{LTL}_f$)~\cite{DeGiacomo2013}.

One of the most important elements in using a formal language in reinforcement learning is the ability to transform a specification into a real-valued function that can be used as reward. This requires quantitative semantics to be defined for the chosen language. One obvious choice is Signal Temporal Logic (STL), which is defined over infinite real-valued signals with a time bound required for every temporal operator. While this is useful for analyzing signals, it can cause problems when defining tasks for robots. For example if the goal is to have the robot learn to put a beer in the fridge, the robot only needs to find the correct way to operate a fridge (e.g. open the fridge door, place the beer on a shelf and close the fridge door) and possibly perform this sequence of actions at an acceptable speed. But using STL to specify this task would require the designer to manually put time bounds on how long each action/subtask should take. If this bound is set inappropriately, the robot may fail to find a satisfying policy due to its hardware constraints even though it is capable of performing the task. This is quite common in robotic tasks where we care about the robot accomplishing the given task but don't have hard constraints on when and how fast the task should be finished. In this case mandatory time bounds add unnecessary complexity to the specification and thus the overall learning process.

Two other possible choices are BLTL and $\textrm{LTL}_f$. Both can be evaluated over finite sequences of states. However, similar to STL, temporal operators in BLTL require time bounds. Both languages are defined over atomic propositions rather than predicates, and do not come with quantitative semantics.

With the above requirements in mind, we design TLTL such that its formulas over state predicates can be evaluated against finite trajectories of any length. In the context of reinforcement learning this can be the length of an execution episode. TLTL does not require a time bound to be specified with every use of a temporal operator. If however the user feels that explicit time bounds are helpful in certain cases, the semantics of STL can be easily incorporated into TLTL. The set of operators provided for TLTL can be conveniently used to specify some common components (goals, constraints, sequences, decisions, etc) that many tasks or rules are made of. The combination of these components can cover a wide range of specifications for robotic tasks.

\section{Related Work}
\label{sec:literature}
Making good use of the reward function in RL has been looked at in the past but has not been the main focus in reinforcement learning research. In \cite{Ng1999}, the authors proposed the method of potential-based reward shaping. It was shown that this method can be used to provide additional training rewards while preserving the optimal policy of the original problem.  Efforts have also been made in inverse reinforcement learning (IRL) where the goal is to extract a reward function from observed optimal/professional behavior, and learn the optimal policy using this reward. Authors in \cite{Ng2000} presented three algorithms that address the problem of IRL and showed their applicability in simple discrete and continuous environments.

Combining temporal logic with reinforcement learning to learn logically complex skills has been looked at only very recently. In \cite{Aksaray2016}, the authors used the log-sum-exp approximation to adapt the robustness of STL specifications to Q-learning on $\tau$-MDPs in discrete spaces. Authors of \cite{Sadigh2014} and~\cite{Fu2014} has also taken advantage of automata-based methods to synthesize control policies that satisfy LTL specifications for MDPs with unknown transition probability.

The methods mentioned above are constrained to discrete state and action spaces, and a somewhat limited set of temporal operators. To the best of our knowledge, this paper is the first to apply TL in reinforcement learning on continuous state and action spaces, and demonstrates its abilities in experimentation.

\section{Experiments}
\label{sec:6}

In this section we first use two simulated manipulation tasks to compare TLTL reward with a discrete reward as well as a distance-based continuous reward commonly used in the RL literature. We then specify a toast placing task in TLTL where a Baxter robot is required to learn a combination of reaching policy and gripper timing policy \footnote{The simulation is implemented in rllab \cite{rllab} and gym \cite{gym}. The experiment is implemented in rllab and ROS}.

\subsection{Simulated 2D Manipulation Tasks}

Figure 2 shows a 2D simulated environment with a three joint manipulator.  The 8 dimensional state feature space includes joint angles, joint velocities and the end-effector position. The 3 dimensional action space includes the joint velocities.  Gaussian noise is added to the velocity commands.

\begin{figure}[tbh]
\label{fig:2}
\begin{center}
\includegraphics[width=1.0\linewidth]{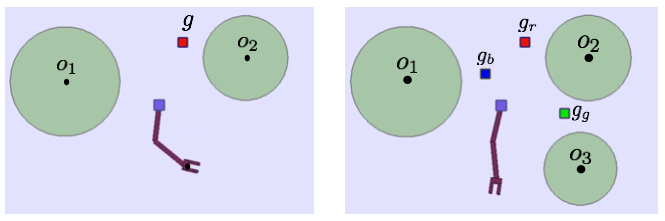}
\caption{2D manipulation tasks. \textit{left}: Task 1. goal reaching while avoiding obstacles. \textit{right}: Task 2. sequential goal reaching while avoiding obstacles}
\end{center}
\end{figure}

\begin{figure*}
\label{fig:3}
\centering
\begin{multicols}{1}
\includegraphics[width=2.\linewidth]{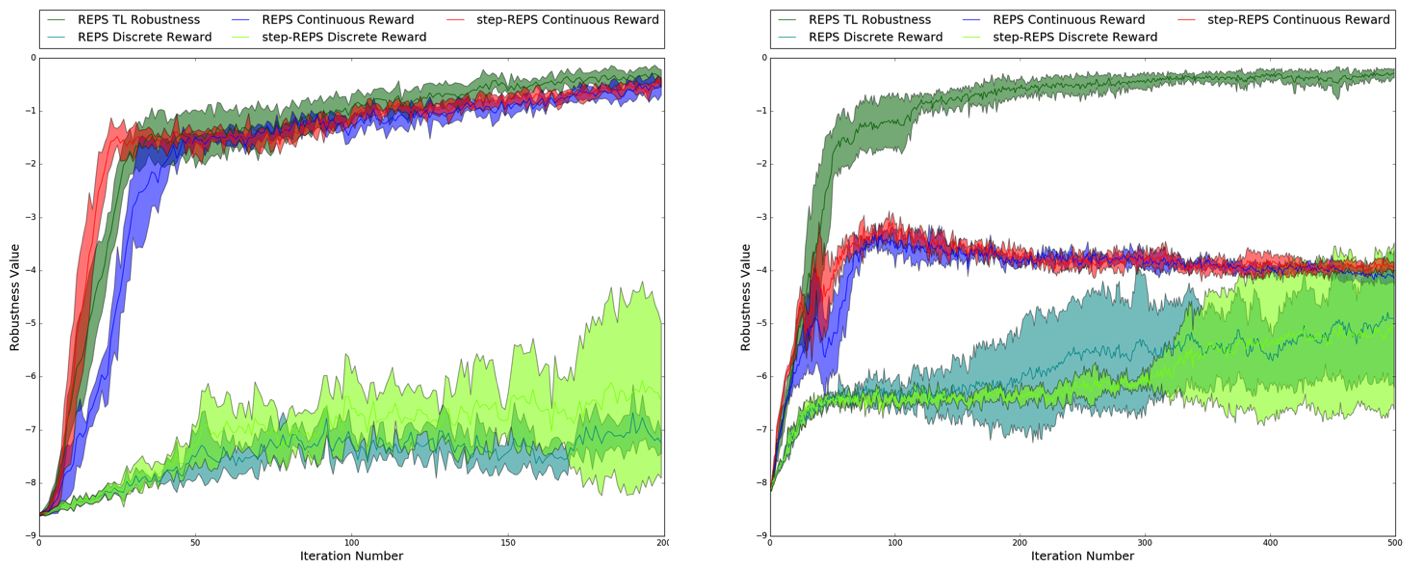}
\end{multicols}
\caption{Learning curves for TLTL robustness, discrete reward and continuous reward trained with episode based REPS, as well as discrete and continuous rewards trained with step based REPS. \textit{left}: task 1, each episode is 200 time-steps, each iteration uses 20 sample trajectories and trained for 200 iterations \textit{right}: task 2, each episode is 500 time-steps, 20 samples per iteration and trained for 500 iterations }
\end{figure*}

For the first task, the end-effector is required to reach the goal position $\boldsymbol g$ while avoiding obstacles $\boldsymbol o_1$ and $\boldsymbol o_2$. The discrete and continuous rewards are summarized as follows:

\begin{equation}
\begin{split}
r_1^{discrete} = \begin{cases}
5 & d_g \leq 0.2 \\
-2 & d_{o_{1,2}} \leq r_{o_{1,2}}\\
0 & \textrm{everywhere else}
\end{cases} \\
r_1^{continuous} =-c_1d_g + c_2\sum_{i=1}^2d_{o_i}.
\end{split}
\end{equation}

\noindent In the above rewards, $d_g$ is the Euclidean distance between the end-effector and the goal, $d_{o_i}$ is the distance between the end-effector and obstacle $i$,$ r_{o_i}$ is the radius of obstacle $i$. The TLTL specification  and its resulting robustness function is described as

\begin{equation}
\phi_1 = \Event\Always(d_g < 0.2) \land \Always(d_{o_1} > r_{o_1} \land d_{o_2} > r_{o_2} )
\end{equation}

\begin{equation}
\begin{split}
\rho_1(\phi_1, (x_e, y_e)_{0:T}) =\min\Bigg(&\underset{t \in [0,T)}{\max}\bigg( \underset{t^\prime \in [t,T)}{\min}\Big( 0.2-d^t_g \Big)\bigg), \\
&\underset{t\in [0,T)}{\min}\Big(d^t_{o_1}-r_{o_1}, d^t_{o_2}-r_{o_2}\Big)\Bigg).
\end{split}
\end{equation}
\noindent In English, $\phi_1$ describes the task of "\textit{eventually always} stay at goal $\boldsymbol g$ \textit{and always} stay alway from obstacles". The user needs only to specify $\phi_1$ and the reward function $\rho_1$ is generated automatically from the quantitative semantics indicated in Section \ref{sec:3a}. Here $(x_e,y_e)_{0:T}$ is the trajectory of the end-effector position. $d^t$ is the is distance at time $t$.

For the second task, the gripper is required to visit goals $\boldsymbol g_r$, $\boldsymbol g_g$, and $\boldsymbol g_b$ in this specific sequence while avoiding the obstacles (one more obstacle is added to further constrain the free space). The discrete and continuous rewards are summarized as

\begin{equation}
\begin{split}
r_2^{discrete} = \begin{cases}
5 & \textrm{goals visited in the right order} \\
-5 & \textrm{goals visited in the wrong}\\
-2 & d_{o_{1,2,3}} \leq r_{o_{1,2,3}}\\
0 & \textrm{everywhere else}
\end{cases} \\
r_2^{continuous} =-c_1d_{g_i} + c_2(d_{g_j}+d_{g_k})+c_3\sum_{i=1}^3d_{o_i}.
\end{split}
\end{equation}

\begin{figure*}
 \centering
 \begin{multicols}{1}
  \includegraphics[width=2.\columnwidth]{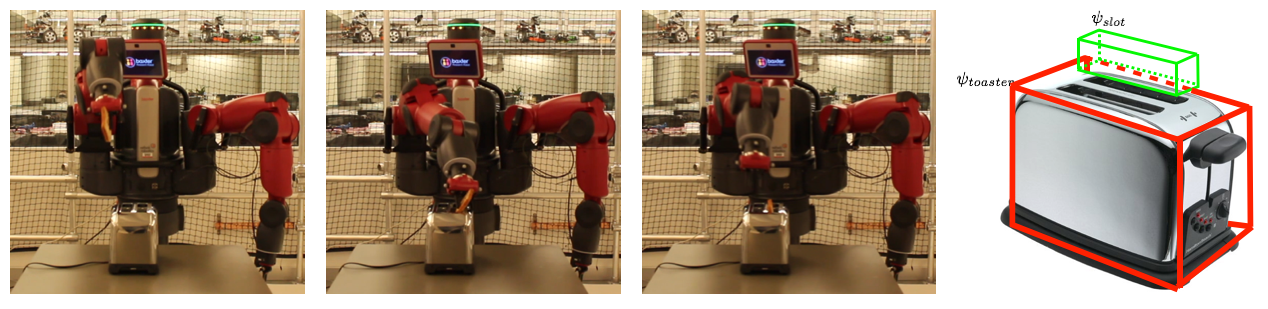}
  \end{multicols}
  \caption{\textit{first three}: Experiment execution. The joint states are measured by encoders, the end-effector states are tracked using the motion tracking system (cameras in the back). \textit{last}: Definition of toaster region predicates}
\end{figure*}

\noindent Here an addition state vector is maintained to record which goals have already been visited in order to know what the next goal is. In $r_2^{continuous}, $ $g_i$ is the correct next goal to visit and $g_j, g_k$ are the goals to avoid. The TLTL specification is defined as

\begin{equation}
\begin{split}
\phi_2=&(\psi_{g_r} \,\, \mathcal{T} \,\, \psi_{g_g} \,\, \mathcal{T} \,\, \psi_{g_b}) \land  (\neg(\psi_{g_g} \lor \psi_{g_b}) \,\, \mathcal{U} \,\, \psi_{g_r}) \land \\
&(\neg(\psi_{g_b}) \,\, \mathcal{U} \,\, \psi_{g_g}) \land (\underset{i=r,g,b}{\bigwedge} \Box(\psi_{g_i} \Rightarrow \bigcirc \Box \neg \psi_{g_i})) \land \Box\psi_o,
\end{split}
\end{equation}
\noindent where $\psi_{g_i}d_{g_i} < 0.2$ is the predicate for goal $g_i$, $\psi_{o}: \underset{j=1,2,3}{\bigwedge}d_{o_j}>r_{o_j}$ is the obstacle avoidance constraint ($\bigwedge$  is a shorthand for a sequence of conjunction). In English, $\phi_2$ states "visit $g_r$ \textit{then} $g_g$ \textit{then} $g_b$, \textit{and don't} visit $g_g$ \textit{or} $g_b$ \textit{until} visiting $g_r$, \textit{and don't} visit $g_b$  \textit{until} visiting $g_g$, \textit{and always} if visited $g_i$ \textit{implies next always don't} visit $g_i$ (don't revisit goals), \textit{and always} avoid obstacles" . Due to space constraints the robustness of $\phi_2$ will not be explicitly presented, but it will also be a complex function consisted of nested $\min()/\max()$ functions that would be difficult to design by hand but can be generated from the quantitative semantics of TLTL.

During training, we consider the obstacles as "penetrable" in that the end-effector of the gripper can enter them with a negative reward received, and depending on the reward function the negative reward may be proportional to the penetration depth. In practice, we find this approach to better facilitate learning than simply granting the agent a negative reward at contact with an obstacle and re-initiate the episode. We will also adopt this approach in the physical experiment in the next section.

Task 1 has a horizon of 200 time-steps, and is trained for 200 iterations with each iteration updated on 30 sample trajectories. Because of the added complexity, task 2 has a horizon of 500 time-steps and is trained for 500 iterations with the same number of samples per update. To compare the influence of reward functions on the learning outcome, we first fix the learning algorithm to be the episode based REPS and compare the average return per iteration for TLTL robustness reward, discrete reward and continuous  reward. However it is meaningless to compare returns on different scales. We therefore take the sample trajectories learned with $r^{discrete}$ and $r^{continuous}$ and calculate their corresponding TLTL robustness return for comparison. The reason for choosing TLTL robustness as the comparison measure is that both the discrete and continuous rewards have semantic ambiguity depending on the choices of the discrete returns and coefficients $c_i$. TLTL is rigorous in its semantics and a robustness greater than zero guarantees satisfaction of the task specification.

In addition, since $r^{discrete}$ and $r^{continuous}$ can provide a immediate reward per step (as oppose to TLTL robustness which requires the entire trajectory to produce a terminal reward), we also used a step based REPS\cite{chebotar2016path} that updates at each step using the cost-to-go. This is a common technique used to reduce the variance in the Monte Carlo return estimate. For continuous rewards, a grid search is performed on the coefficients $c_i$ and the best outcome is reported. We train each comparison case on 4 different random seeds. The mean and variance of the average returns are illustrated in Figure 3 .

It can be observed that in both tasks TLTL robustness reward resulted in the best learning outcome in terms of convergence rate and final return.  For the level of stochasticity presented in the simulation, step based REPS showed only minor improvement in the rate of convergence and variance reduction. For the simpler case of task 1, a well tuned continuous reward achieves comparable learning performance with the TLTL robustness reward. For task 2, the TLTL reward outperforms competing reward functions by a considerable margin. Discrete reward fails to learn a useful policy due to sparse returns. A video of the learning process is provided.

The results indicate that a reward function with well defined semantics can significantly improve the learning outcome of an agent while providing convenience to the designer. For tasks with a temporal/causal structure (such as task 2), a hierarchical learning approach is usually employed where the agent learns higher level policies that schedules over lower level ones\cite{dietterich2000hierarchical}. We show that incorporating the temporal structure correctly into the reward function allows for a relatively simple non-hierarchical algorithm to learn hierarchical tasks in continuous state and action spaces.

\subsection{Learning Toast-Placing Task With A Baxter Robot}

Pick-and-placement tasks have been a common test scenario in reinforcement learning research \cite{Levine2016},\cite{gu2016deep}. The task is framed as correctly reaching a grasp position where the end-effector will perform the grasp operation upon approach. For the object placing process, progress is measured by  tracking the distance between the object and the place to deploy. In our experiment, we will be focusing on the placing task. We will not be tracking the position of the object but rather express the desired behavior as a TLTL specification. The robot will simultaneously learn to reach the specified region and a gripper timing policy that releases the object at the right instant (as oppose to directly specifying the point of release).

Figure 4 shows the experimental setup. A Baxter robot is used to perform the task of placing a piece of bread in a toaster. The 21 dimensional state feature space includes 7 joint angles and joint velocities, the xyz-rpy pose of the end-effector and the gripper position. The end-effector pose is tracked using the motion tracking system as an additional source of information. The gripper position ranges continuously from 0 to 100 with 0 being fully closed. The 8 dimensional action space includes 7 joint velocities and the desired gripper position. Actions are sent at 20hz.

The placing task is specified by the TLTL formula

\begin{equation}
\begin{split}
\phi=&\Box(\neg(\psi_{table} \vee \psi_{toaster})) \wedge \diamondsuit(\psi_{slot}) \wedge \\
&(\psi_{gc} \,\, \mathcal{U} \,\, \psi_{slot}) \wedge \Box(\psi_{slot} \Rightarrow \bigcirc\Box(\psi_{go})) ,
\end{split}
\end{equation}

\noindent where $\psi_{table}$, $\psi_{toaster}$, $\psi_{slot}$ are predicates describing spatial regions in the form $(x_{min} < x_e < x_{max}) \wedge (y_{min} < y_e < y_{max}) \wedge (z_{min} < z_e < z_{max})$ ($(x_e,y_e,z_e)$ is the position of the end-effector). Orientation constraints are specified in a similar way to ensure the correct pose is reached at the position of release. The regions for $slot$ and $toaster$ are depicted in Figure 4. $\psi_{gc}: p_g < \delta_{close}$ and $\psi_{go}: p_g > \delta_{open}$
describe the conditions for gripper open/close.  In English, the specification describes the process of "\textit{always don't} hit the table \textit{or} the toaster, \textit{and eventually} reach the slot, \textit{and} keep gripper closed \textit{until} slot is reached, \textit{and always} if slot is reached \textit{implies next always}  keep gripper open". The resulting robustness for $\phi$ is

\begin{equation}
\label{eq:15}
\begin{split}
&\rho(\phi,p^e_{0:T}) = \\
&\min\Bigg( \underset{t \in [0,T)}{\min} \bigg(  \max\Big( -\rho(\psi_{table}, p^e_{t:T}), -\rho(\psi_{toaster}, p^e_{t:T}) \Big)\bigg), \\
& \underset{t \in [0,T)}{\max}\rho(\psi_{slot}, p^e_{t:T}), \\
&\underset{t \in [0,T)}{\max} \bigg(  \min\Big( \rho(\psi_{slot}, p^e_{t:T}), \underset{t^\prime \in [0,t)}{\min}\rho(\psi_{gc}, p^e_{t^\prime:t}) \Big)\bigg),\\
&\underset{t \in [0,T)}{\min} \bigg(  \max\Big( -\rho(\psi_{slot}, p^e_{t:T}), \underset{t^\prime \in [t+1,T)}{\min}\rho(\psi_{go}, p^e_{t^\prime:T}) \Big)\bigg)\Bigg).
\end{split}
\end{equation}
 \noindent Due to space constraints~\eqref{eq:15} is written in its recursive form where the robustness of the individual predicates are evaluated at each time step. Again $\rho(\phi,p^e_{0:T})$ is generated from the TLTL quantitative semantics and specification $\phi$ is satisfied when $\rho(\phi,p^e_{0:T}) > 0$ . Also the robustness for $\psi_{gc}$ and $\psi_{go}$ are normalized to the same scale as that of the other predicates. This is to ensure that all sub-formulas are treated equally during learning.  Implementation of~\eqref{eq:15} or robustness in general is highly vectorized and run time evaluation speed for complicated specifications do not usually cause significant overhead.

For a comparison case, we design the following reward function

 \begin{equation}
 \begin{split}
 &r_t =\\
 &\begin{cases}
      -c_1d^t_{slot} + c_2d^t_{toaster} - c_3|p^t_g| & \underset{t^\prime \in [0,t)}{\min}d^t_{slot} > 0.03 \\
      -c_1d^t_{slot} + c_2d^t_{toaster} - c_3|100-p^t_g| & \underset{t^\prime \in [0,t)}{\min}d^t_{slot} < 0.03.
   \end{cases}
 \end{split}
 \end{equation}

 \noindent In the above equation, $d^t_{slot}$ and $d^t_{toaster}$ are the Euclidean distances between the end-effector and the center of the toaster regions defined in Figure 4 (at time $t$). $p^t_g$ is the gripper position at time $t$. The reward function encourages being close to the slot and keeping away from the toaster. If the gripper has yet to reached to within 3 centimeters of the slot center at all times before $t$, the gripper should be closed ($p^t_g = 0$), otherwise the gripper should be opened. The coefficients $c_{1,2,3}$ are manually tuned and the best outcome is reported.

 Similar to the simulation experiment, during training the obstacle (toaster in this case) is taken away and the region $\psi_{toaster}$ is penetrable with a negative reward proportional to the penetration depth (highest at the center of the region) provided by the robustness. The table is kept at its position and a new episode is initialized if collision with the table occurs. Each episode has a horizon of 100 time-steps (around 6 seconds) and each update iteration uses 10 sample trajectories. Episode based REPS is again used as the RL algorithm for this task. The resulting training curves are plotted in Figure 5.

 \begin{figure}[tbh]
\label{fig:5}
\begin{center}
\includegraphics[width=1.05\linewidth]{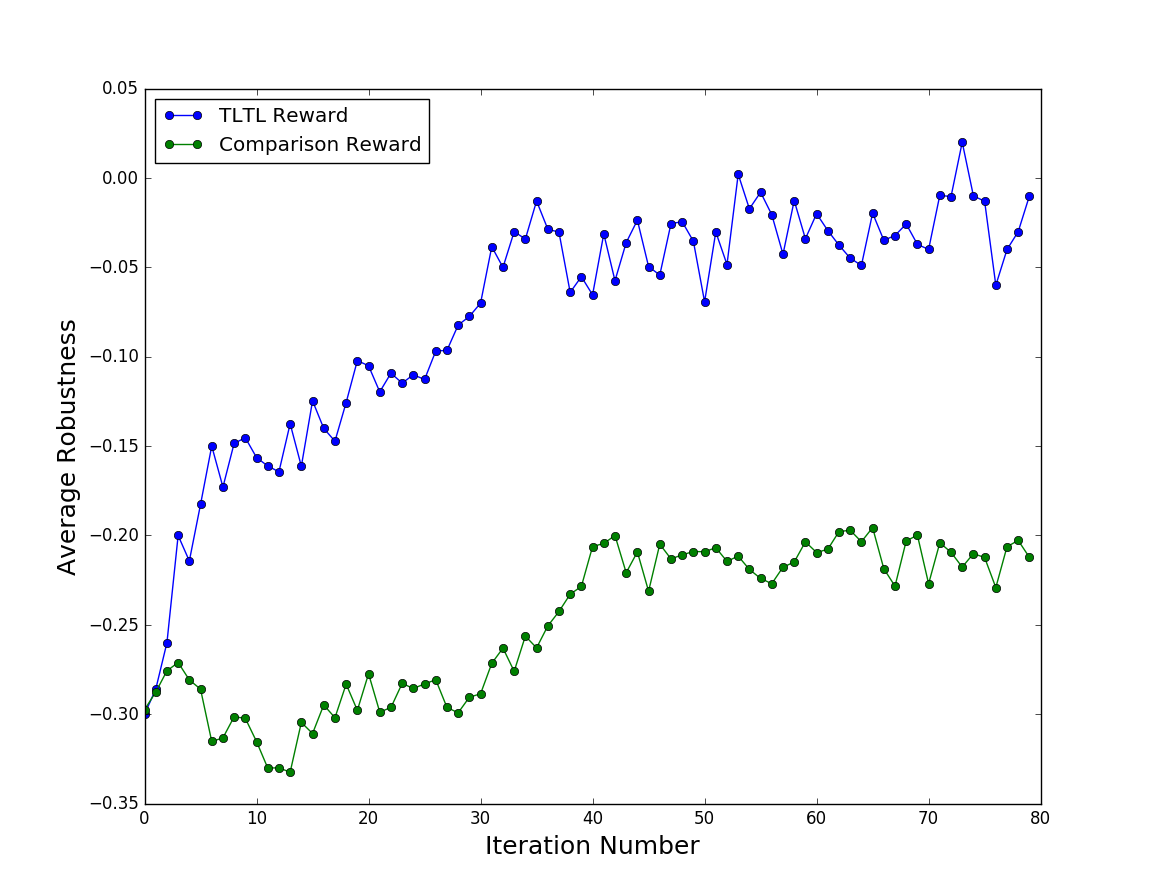}
\caption{Training curves for Baxter toast-placing task. An episode is 100 time-steps long (around 6 seconds). Each update iteration uses 10 sample trajectories. Trained for 80 iterations}
\end{center}
\end{figure}

In Figure 5, trajectories learned from $r_t$ at each iteration are used to calculated their corresponding robustness value (as explained in the previous section) for a reasonable comparison. We can observe that training with TLTL reward has reached a significantly better policy than that with the comparison reward. One important reason is that the semantics of $r_t$ in Equation (16) relies heavily on the relative magnitudes of the coefficients $c_{1,2,3}$. For example if $c_1$ is much higher than $c_2$ and $c_3$, then $r_t$ will put most emphasis on reaching the slot and pay less attention on learning the correct gripper timing policy or obstacle avoidance. An exhaustive hyperparameter search on the physical robot is infeasible. In addition, $r_t$ expresses much less information than $\rho(\phi,p^e_{0:T})$. For example, penalizing collision with the toaster is necessary only when the gripper comes in contact with the toaster. Otherwise the agent should focus on the other subtasks (reaching the slot, improving the gripper policy). For reward $r_t$, this logistics is again achieved only by obtaining the right combination of hyperparameters.  However, because the robustness function is made up of a series of embedded $\min()/\max()$ functions, at any instant the agent will be maximizing only a set of active functions. These active functions represent the bottlenecks in improving the overall return. By adopting this form, the robustness reward effectively focuses the agent's effort in improving the most critical set of subtasks at any time so to achieve an efficient overall learning progress. However, this may render the TLTL robustness reward susceptible to scaling (if the robustness of a sub-formulae changes on a different scale than other sub-formula, the agent may devote all its effort in improving on this one sub-task and fail to improve on the others). Therefore, proper normalization is required. Currently this normalization process is achieved manually, future work can include automatic or adaptive normalization of predicate robustnesses.

To evaluate the resulting behavior, 10 trials of the toast-placing task is executed with the policy learned from each reward. The policy from the TLTL reward achieves 100\% success rate while the comparison reward fails to learn the task (due to its inability to learn the correct gripper time policy). A video of the learning progress is provided.

\section{Conclusion}
\label{sec:conclusion}

Looking over our learning process as humans, we are usually given a goal and a set of well defined rules. And it is up to us to find methods to best achieve the goal within the given rules. But imagine if we are only given the goal (drive safely from A to B) but not the rules (traffic rules), even if we can instantly reinitialize after the occurrence of an accident it would take an intractable number of trials before we learn to drive safely. Robot learning is analogous. Rules can be experience/knowledge (switch to low gear if driving on steep slopes) that accelerate learning, or they can be constraints that an agent must abide by (rules in traffic, sports and games). Being able to formally express these rules as reward functions and incorporate them in the learning process is helpful and often necessary for a robot to operate in the world.

In this paper we proposed TLTL, a formal specification language with quantitative semantics that is designed for convenient robotic task specification. We compare learning performance of the TLTL reward with two of the more commonly used forms of reward (namely a discrete and continuous form of reward functions) in a 2D simulated manipulation environment by fixing the RL algorithm. We also compare the outcome of TLTL reward trained using a relatively inefficient episode based method with the discrete/continuous rewards trained using a lower variance step based method. Results show that TLTL reward not only outperformed all of its comparison cases, it also enabled a non-hierarchical RL method to successfully learn to perform a temporally structured task. Furthermore, We used TLTL to express a toast-placing task and demonstrated successful learning on a Baxter robot. Future work includes adapting TLTL robustness reward to more efficient
gradient based methods \cite{heess2015memory} by exploiting its smooth approximations \cite{Aksaray2016}. And using automata theory for guided exploration and learning \cite{aydin2012language}.

%% Use plainnat to work nicely with natbib.

\bibliographystyle{IEEEtran}
\bibliography{references.bib}

\end{document}